# Two-Level, Many-Paths Generation

**Kevin Knight**

USC/Information Sciences Institute
4676 Admiralty Way
Marina del Rey, CA 90292
knight@isi.edu

**Vasileios Hatzivassiloglou**

Department of Computer Science
Columbia University
New York, NY 10027
vh@cs.columbia.edu

## Abstract

Large-scale natural language generation requires the integration of vast amounts of knowledge: lexical, grammatical, and conceptual. A robust generator must be able to operate well even when pieces of knowledge are missing. It must also be robust against incomplete or inaccurate inputs. To attack these problems, we have built a hybrid generator, in which gaps in symbolic knowledge are filled by statistical methods. We describe algorithms and show experimental results. We also discuss how the hybrid generation model can be used to simplify current generators and enhance their portability, even when perfect knowledge is in principle obtainable.

## 1 Introduction

A large-scale natural language generation (NLG) system for unrestricted text should be able to operate in an environment of 50,000 conceptual terms and 100,000 words or phrases. Turning conceptual expressions into English requires the integration of large knowledge bases (KBs), including grammar, ontology, lexicon, collocations, and mappings between them. The quality of an NLG system depends on the quality of its inputs and knowledge bases. Given that perfect KBs do not yet exist, an important question arises: can we build high-quality NLG systems that are robust against incomplete KBs and inputs? Although robustness has been heavily studied in natural language understanding (Weischedel and Black, 1980; Hayes, 1981; Lavie, 1994), it has received much less attention in NLG (Robin, 1995).

We describe a hybrid model for natural language generation which offers improved performance in the presence of knowledge gaps in the generator (the grammar and the lexicon), and of errors in the semantic input. The model comes out of our practical experience in building a large Japanese-English newspaper machine translation system, JAPANGLOSS (Knight et al., 1994; Knight et al., 1995). This system translates Japanese into representations whose terms are drawn from the SENSUS ontology (Knight and Luk, 1994), a 70,000-node knowledge base skeleton derived from resources like WordNet (Miller, 1990), Longman's Dictionary (Procter, 1978), and the PENMAN Upper Model (Bateman, 1990). These representations are turned into English during generation. Because we are processing unrestricted newspaper text, all modules in JAPANGLOSS must be robust.

In addition, we show how the model is useful in simplifying the design of a generator and its knowledge bases even when perfect knowledge is available. This is accomplished by relegating some aspects of lexical choice (such as preposition selection and non-compositional interlexical constraints) to a statistical component. The generator can then use simpler rules and combine them more freely; the price of this simplicity is that some of the output may be invalid. At this point, the statistical component intervenes and filters from the output all but the fluent expressions. The advantage of this two-level approach is that the knowledge bases in the generator become simpler, easier to develop, more portable across domains, and more accurate and robust in the presence of knowledge gaps.

## 2 Knowledge Gaps

In our machine translation experiences, we traced generation disfluencies to two sources:[1] (1) incomplete or inaccurate conceptual (interlingua) structures, caused by knowledge gaps in the source language analyzer, and (2) knowledge gaps in the generator itself. These two categories of gaps include:

- Interlingual analysis often does not include accurate representations of number, definiteness, or time. (These are often unmarked in Japanese and require exceedingly difficult inferences to recover).

- The generation lexicon does not mark rare words and generally does not distinguish between near synonyms (e.g., *finger* vs. ?*digit*).

---

[1] See also (Kukich, 1988) for a discussion of fluency problems in NLG systems.

- The generation lexicon does not contain much collocational knowledge (e.g., *on the field* vs. \**on the end zone*).
- Lexico-syntactic constraints (e.g., *tell her hi* vs. \**say her hi*), syntax-semantics mappings (e.g., *the vase broke* vs. \**the food ate*), and selectional restrictions are not always available or accurate.

The generation system we use, PENMAN (Penman, 1989), is robust because it supplies appropriate defaults when knowledge is missing. But the default choices frequently are not the optimal ones; the hybrid model we describe provides more satisfactory solutions.

## 3 Issues in Lexical Choice

The process of selecting words that will lexicalize each semantic concept is intrinsically linked with syntactic, semantic, and discourse structure issues.[2] Multiple constraints apply to each lexical decision, often in a highly interdependent manner. However, while some lexical decisions can affect future (or past) lexical decisions, others are purely local, in the sense that they do not affect the lexicalization of other semantic roles. Consider the case of time adjuncts that express a single point in time, and assume that the generator has already decided to use a prepositional phrase for one of them. There are several forms of such adjuncts, e.g.,

$$\text{She left} \begin{cases} \text{at five.} \\ \text{on Monday.} \\ \text{in February.} \end{cases}$$

In terms of their interactions with the rest of the sentence, these manifestations of the adjunct are identical. The use of different prepositions is an interlexical constraint between the semantic and syntactic heads of the PP that does not propagate outside the PP. Consequently, the selection of the preposition can be postponed until the very end.

Existing generation models however select the preposition according to defaults or randomly among possible alternatives or by explicitly encoding the lexical constraints. The PENMAN generation system (Penman, 1989) defaults the preposition choice for point-time adjuncts to *at*, the most commonly used preposition in such cases. The FUF/SURGE (Elhadad, 1993) generation system is an example where prepositional lexical restrictions in time adjuncts are encoded by hand, producing fluent expressions but at the cost of a larger grammar.

Collocational restrictions are another example of lexical constraints. Phrases such as *three straight victories*, which are frequently used in sports reports to express historical information, can be decomposed semantically into the head noun plus its modifiers. However, when ellipsis of the head noun is considered, a detailed corpus analysis of actual basketball game reports (Robin, 1995) shows that the forms *won/lost three straight X*, *won/lost three consecutive X*, and *won/lost three straight* are regularly used, but the form \**won/lost three consecutive* is not. To achieve fluent output within the knowledge-based generation paradigm, lexical constraints of this type must be explicitly identified and represented.

Both the above examples indicate the presence of (perhaps domain-dependent) lexical constraints that are not explainable on semantic grounds. In the case of prepositions in time adjuncts, the constraints are institutionalized in the language, but still nothing about the concept MONTH relates to the use of the preposition *in* with month names instead of, say, *on* (Herskovits, 1986). Furthermore, lexical constraints are not limited to the syntagmatic, interlexical constraints discussed above. For a generator to be able to produce sufficiently varied text, multiple renditions of the same concept must be accessible. Then, the generator is faced with paradigmatic choices among alternatives that without sufficient information may look equivalent. These choices include choices among synonyms (and near-synonyms), and choices among alternative syntactic realizations of a semantic role. However, it is possible that not all the alternatives actually share the same level of fluency or currency in the domain, even if they are rough paraphrases.

In short, knowledge-based generators are faced with multiple, complex, and interacting lexical constraints,[3] and the integration of these constraints is a difficult problem, to the extent that the need for a different specialized architecture for lexical choice in each domain has been suggested (Danlos, 1986). However, compositional approaches to lexical choice have been successful whenever detailed representations of lexical constraints can be collected and entered into the lexicon (e.g., (Elhadad, 1993; Kukich et al., 1994)). Unfortunately, most of these constraints must be identified manually, and even when automatic methods for the acquisition of some types of this lexical knowledge exist (Smadja and McKeown, 1991), the extracted constraints must still be transformed to the generator's representation language by hand. This narrows the scope of the lexicon to a specific domain; the approach fails to scale up to unrestricted language. When the goal is domain-independent generation, we need to investigate methods for producing reasonable output *in the absence of* a large part of the information tradi-tionally available to the lexical chooser.

---

[2] We consider lexical choice as a general problem for both open and closed class words, not limiting it to the former only as is sometimes done in the generation literature.

[3] Including constraints not discussed above, originating for example from discourse structure, the user models for the speaker and hearer, and pragmatic needs.

## 4 Current Solutions

Two strategies have been used in lexical choice when knowledge gaps exist: selection of a default,[4] and random choice among alternatives. Default choices have the advantage that they can be carefully chosen to mask knowledge gaps to some extent. For example, PENMAN defaults article selection to *the* and tense to present, so it will produce *The dog chases the cat* in the absence of definiteness information. Choosing *the* is a good tactic, because *the* works with mass, count, singular, plural, and occasionally even proper nouns, while *a* does not. On the down side, *the*'s only outnumber *a*'s and *an*'s by about two-to-one (Knight and Chander, 1994), so guessing *the* will frequently be wrong. Another ploy is to give preference to nominalizations over clauses. This generates sentences like *They plan the statement of the filing for bankruptcy*, avoiding disasters like *They plan that it is said to file for bankruptcy*. Of course, we also miss out on sparkling renditions like *They plan to say that they will file for bankruptcy*. The alternative of randomized decisions offers increased paraphrasing power but also the risk of producing some non-fluent expressions; we could generate sentences like *The dog chased a cat* and *A dog will chase the cat*, but also *An earth circles a sun*.

To sum up, defaults can help against knowledge gaps, but they take time to construct, limit paraphrasing power, and only return a mediocre level of quality. We seek methods that can do better.

## 5 Statistical Methods

Another approach to the problem of incomplete knowledge is the following. Suppose that according to our knowledge bases, input $I$ may be rendered as sentence $A$ or sentence $B$. If we had a device that could invoke new, easily obtainable knowledge to score the input/output pair $\langle I, A \rangle$ against $\langle I, B \rangle$, we could then choose $A$ over $B$, or vice-versa. An alternative to this is to forget $I$ and simply score $A$ and $B$ on the basis of fluency. This essentially assumes that our generator produces valid mappings from $I$, but may be unsure as to which is the correct rendition. At this point, we can make another approximation—modeling fluency as likelihood. In other words, how often have we seen $A$ and $B$ in the past? If $A$ has occurred fifty times and $B$ none at all, then we choose $A$. But if $A$ and $B$ are long sentences, then probably we have seen neither. In that case, further approximations are required. For example, does $A$ contain frequent three-word sequences? Does $B$?

Following this reasoning, we are led into statistical language modeling. We built a language model for the English language by estimating bigram and trigram probabilities from a large collection of 46 million words of Wall Street Journal material.[5] We smoothed these estimates according to class membership for proper names and numbers, and according to an extended version of the *enhanced Good-Turing method* (Church and Gale, 1991) for the remaining words. The latter smoothing operation not only optimally regresses the probabilities of seen n-grams but also assigns a non-zero probability to all unseen n-grams which depends on how likely their component m-grams ($m < n$, i.e., words and bigrams) are. The resulting conditional probabilities are converted to log-likelihoods for reasons of numerical accuracy and used to estimate the overall probability $P(S)$ of any English sentence $S$ according to a Markov assumption, i.e.,

$$\log P(S) = \sum_i \log P(w_i | w_{i-1}) \quad \text{for bigrams}$$

$$\log P(S) = \sum_i \log P(w_i | w_{i-1}, w_{i-2}) \quad \text{for trigrams}$$

Because both equations would assign lower and lower probabilities to longer sentences and we need to compare sentences of different lengths, a heuristic strictly increasing function of sentence length, $f(l) = 0.5l$, is added to the log-likelihood estimates.

## 6 First Experiment

Our first goal was to integrate the symbolic knowledge in the PENMAN system with the statistical knowledge in our language model. We took a semantic representation generated automatically from a short Japanese sentence. We then used PENMAN to generate 3,456 English sentences corresponding to the 3,456 ($= 2^7 \cdot 3^3$) possible combinations of the values of seven binary and three ternary features that were unspecified in the semantic input. These features were relevant to the semantic representation but their values were not extractable from the Japanese sentence, and thus each of their combinations corresponded to a particular interpretation among the many possible in the presence of incompleteness in the semantic input. Specifying a feature forced PENMAN to make a particular linguistic decision. For example, adding (`:identifiability-q t`) forces the choice of determiner, while the `:lex` feature offers explicit control over the selection of open-class words. A literal translation of the input sentence was something like *As for new company, there is plan to establish in February*. Here are three randomly selected translations; note that the object of the "establishing" action is unspecified in the Japanese input, but PENMAN supplies a placeholder *it* when necessary, to ensure grammaticality:

---

[4]See also (Harbusch et al., 1994) for a thorough discussion of defaulting in NLG systems.

[5]Available from the ACL Data Collection Initiative, as CD ROM 1.

```
   A new company will have in mind that it
      is establishing it on February.

   The new company plans the launching
      on February.

   New companies will have as a goal
      the launching at February.
```

We then ranked the 3,456 sentences using the bigram version of our statistical language model, with the hope that good renditions would come out on top. Here is an abridged list of outputs, log-likelihood scores heuristically corrected for length, and rankings:

```
  1  The new company plans to
        launch it in February.    [ -13.568260 ]
  2  The new company plans the
        foundation in February.   [ -13.755152 ]
  3  The new company plans the
        establishment in February. [ -13.821412 ]
  4  The new company plans to
        establish it in February. [ -14.121367 ]
     ...................................
 60  The new companies plan the
        establishment on February. [ -16.350112 ]
 61  The new companies plan the
        launching in February.    [ -16.530286 ]
     ...................................
400  The new companies have as a goal the
        foundation at February.   [ -23.836556 ]
401  The new companies will have in mind to
        establish it at February. [ -23.842337 ]
     ...................................
```

While this experiment shows that statistical models can help make choices in generation, it fails as a computational strategy. Running PENMAN 3,456 times is expensive, but nothing compared to the cost of exhaustively exploring all combinations in larger input representations corresponding to sentences typically found in newspaper text. Twenty or thirty choice points typically multiply into millions or billions of potential sentences, and it is infeasible to generate them all independently. This leads us to consider other algorithms.

## 7 Many-Paths Generation

Instead of explicitly constructing all possible renditions of a semantic input and running PENMAN on them, we use a more efficient data structure and control algorithm to express possible ambiguities. The data structure is a *word lattice*—an acyclic state transition network with one start state, one final state, and transitions labeled by words. Word lattices are commonly used to model uncertainty in speech recognition (Waibel and Lee, 1990) and are well adapted for use with n-gram models.

As we discussed in Section 3, a number of generation difficulties can be traced to the existence of constraints between words and phrases. Our generator operates on *lexical islands*, which do not interact with other words or concepts.[6] How to identify such islands is an important problem in NLG: grammatical rules (e.g., agreement) may help group words together, and collocational knowledge can also mark the boundaries of some lexical islands (e.g., nominal compounds). When no explicit information is present, we can resort to treating single words as lexical islands, essentially adopting a view of maximum compositionality. Then, we rely on the statistical model to correct this approximation, by identifying any violations of the compositionality principle on the fly during actual text generation.

The type of the lexical islands and the manner by which they have been identified do not affect the way our generator processes them. Each island corresponds to an independent component of the final sentence. Each individual word in an island specifies a choice point in the search and causes the creation of a state in the lattice; all continuations of alternative lexicalizations for this island become paths that leave this state. Choices between alternative lexical islands for the same concept also become states in the lattice, with arcs leading to the sub-lattices corresponding to each island.

Once the semantic input to the generator has been transformed to a word lattice, a search component identifies the $N$ highest scoring paths from the start to the final state, according to our statistical language model. We use a version of the N-best algorithm (Chow and Schwartz, 1989), a Viterbi-style beam search algorithm that allows extraction of more than just the best scoring path. (Hatzivassiloglou and Knight, 1995) has more details on our search algorithm and the method we applied to estimate the parameters of the statistical model.

Our approach differs from traditional top-down generation in the same way that top-down and bottom-up parsing differ. In top-down parsing, backtracking is employed to exhaustively examine the space of possible alternatives. Similarly, traditional control mechanisms in generation operate top-down, either deterministically (Meteer et al., 1987; Tomita and Nyberg, 1988; Penman, 1989) or by backtracking to previous choice points (Elhadad, 1993). This mode of operation can unnecessarily duplicate a lot of work at run time, unless sophisticated control directives are included in the search engine (Elhadad and Robin, 1992). In contrast, in bottom-up parsing and in our generation model, a special data structure (a chart or a lattice respectively) is used to efficiently encode multiple analyses, and to allow structure sharing between many alternatives, eliminating repeated search.

What should the word lattices produced by a generator look like? If the generator has complete

---

[6]At least as far as the generator knows.

knowledge, the word lattice will degenerate to a string, e.g.:

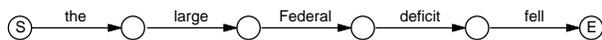

Suppose we are uncertain about definiteness and number. We can generate a lattice with eight paths instead of one:

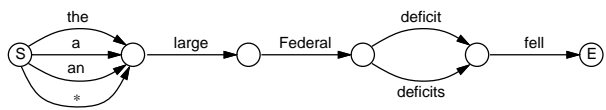

(* stands for the empty string.) But we run the risk that the n-gram model will pick a non-grammatical path like *a large Federal deficits fell*. So we can produce the following lattice instead:

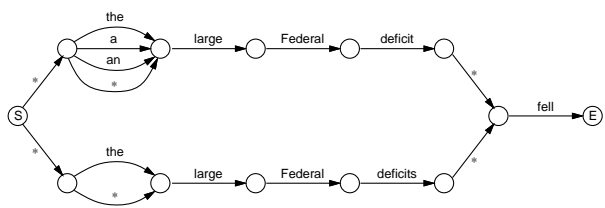

In this case, we use knowledge about agreement to constrain the choices offered to the statistical model, from eight paths down to six. Notice that the six-path lattice has more states and is more complex than the eight-path one. Also, the n-gram length is critical. When long-distance features control grammaticality, we cannot rely on the statistical model. Fortunately, long-distance features like agreement are among the first that go into any symbolic generator. This is our first example of how symbolic and statistical knowledge sources contain complementary information, which is why there is a significant advantage to combining them.

Now we need an algorithm for converting generator inputs into word lattices. Our approach is to assign word lattices to each fragment of the input, in a bottom-up compositional fashion. For example, consider the following semantic input, which is written in the PENMAN-style Sentence Plan Language (SPL) (Penman, 1989), with concepts drawn from the SENSUS ontology (Knight and Luk, 1994), and may be rendered in English as *It is easy for Americans to obtain guns*:

```
(A / |have the quality of being|
  :DOMAIN (P / |procure|
             :AGENT (A2 / |American|)
             :PATIENT (G / |gun, arm|))
  :RANGE (E / |easy, effortless|))
```

We process semantic subexpressions in a bottom-up order, e.g., `A2`, `G`, `P`, `E`, and finally `A`. The grammar assigns what we call an *e-structure* to each subexpression. An e-structure consists of a list of distinct syntactic categories, paired with English word lattices: (<syn, lat>, <syn, lat>, ...). As we climb up the input expression, the grammar glues together various word lattices. The grammar is organized around semantic feature patterns rather than English syntax—rather than having one `S -> NP-VP` rule with many semantic triggers, we have one `AGENT-PATIENT` rule with many English renderings. Here is a sample rule:

```
((x1 :agent) (x2 :patient) (x3 :rest)
->
(s (seq (x1 np) (x3 v-tensed) (x2 np)))
(s (seq (x1 np) (x3 v-tensed) (wrd "that")
        (x2 s)))
(s (seq (x1 np) (x3 v-tensed)
        (x2 (*OR* inf inf-raise))))
(s (seq (x2 np) (x3 v-passive) (wrd "by")
        (x1 np)))
(inf (seq (wrd "for") (x1 np) (wrd "to")
          (x3 v) (x2 np)))
(inf-raise (seq (x1 np)
                (or (seq (wrd "of") (x3 np)
                         (x2 np))
                    (seq (wrd "to") (x3 v)
                         (x2 np)))))
(np (seq (x3 np) (wrd "of") (x2 np)
         (wrd "by") (x1 np))))
```

Given an input semantic pattern, we locate the first grammar rule that matches it, i.e., a rule whose left-hand-side features except `:rest` are contained in the input pattern. The feature `:rest` is our mechanism for allowing partial matchings between rules and semantic inputs. Any input features that are not matched by the selected rule are collected in `:rest`, and recursively matched against other grammar rules.

For the remaining features, we compute new e-structures using the rule's right-hand side. In this example, the rule gives four ways to make a syntactic `S`, two ways to make an infinitive, and one way to make an `NP`. Corresponding word lattices are built out of elements that include:

- `(seq x y ...)`—create a lattice by sequentially gluing together the lattices `x`, `y`, and ...
- `(or x y ...)`—create a lattice by branching on `x`, `y`, and ...
- `(wrd w)`—create the smallest lattice: a single arc labeled with the word `w`.
- `(xn <syn>)`—if the e-structure for the semantic material under the `xn` feature contains <syn, lat>, return the word lattice `lat`; otherwise fail.

Any failure inside an alternative right-hand side of a rule causes that alternative to fail and be ignored. When all alternatives have been processed, results are collected into a new e-structure. If two or more word lattices can be created from one rule, they are merged with a final `or`.

Because our grammar is organized around semantic patterns, it nicely concentrates all of the material required to build word lattices. Unfortunately, it forces us to restate the same syntactic constraint in many places. A second problem is that sequential composition does not allow us to insert new words inside old lattices, as needed to generate sentences like *John looked it up.* We have extended our notation to allow such constructions, but the full solution is to move to a unification-based framework, in which e-structures are replaced by arbitrary feature structures with `syn`, `sem`, and `lat` fields. Of course, this requires extremely efficient handling of the disjunctions inherent in large word lattices.

## 8  Results

We implemented a medium-sized grammar of English based on the ideas of the previous section, for use in experiments and in the JAPANGLOSS machine translation system. The system converts a semantic input into a word lattice, sending the result to one of three sentence extraction programs:

- `RANDOM`—follows a random path through the lattice.
- `DEFAULT`—follows the topmost path in the lattice. All alternatives are ordered by the grammar writer, so that the topmost lattice path corresponds to various defaults. In our grammar, defaults include singular noun phrases, the definite article, nominal direct objects, *in* versus *on*, active voice, *that* versus *who*, the alphabetically first synonym for open-class words, etc.
- `STATISTICAL`—a sentence extractor based on word bigram probabilities, as described in Sections 5 and 7.

For evaluation, we compare English outputs from these three sources. We also look at lattice properties and execution speed. Space limitations prevent us from tracing the generation of many long sentences—we show instead a few short ones. Note that the sample sentences shown for the `RANDOM` extraction model are not of the quality that would normally be expected from a knowledge-based generator, because of the high degree of ambiguity (unspecified features) in our semantic input. This incompleteness can be in turn attributed in part to the lack of such information in Japanese source text and in part to our own desire to find out how much of the ambiguity can be automatically resolved with our statistical model.

---

INPUT

```
(A / |accuse|
  :AGENT SHE
  :PATIENT (T / |thieve|
             :AGENT HE
             :PATIENT (M / |motorcar|)))
```

LATTICE CREATED

  44 nodes, 217 arcs, 381,440 paths;
  59 distinct unigrams, 430 distinct bigrams.

RANDOM EXTRACTION

  Her incriminates for him to thieve an
  automobiles.

  She am accusing for him to steal autos.

  She impeach that him thieve that there
  was the auto.

DEFAULT EXTRACTION

  She accuses that he steals the auto.

STATISTICAL BIGRAM EXTRACTION

  1  She charged that he stole the car.
  2  She charged that he stole the cars.
  3  She charged that he stole cars.
  4  She charged that he stole car.
  5  She charges that he stole the car.

TOTAL EXECUTION TIME:  22.77 CPU seconds.

---

INPUT

```
(A / |have the quality of being|
  :DOMAIN (P / |procure|
            :AGENT (A2 / |American|)
            :PATIENT (G / |gun, arm|))
  :RANGE (E / |easy, effortless|))
```

LATTICE CREATED

  64 nodes, 229 arcs, 1,345,536 paths;
  47 distinct unigrams, 336 distinct bigrams.

RANDOM EXTRACTION

  Procurals of guns by Americans were easiness.

  A procurements of guns by a Americans will
  be an effortlessness.

  It is easy that Americans procure that
  there is gun.

DEFAULT EXTRACTION

  The procural of the gun by the American is
  easy.

STATISTICAL BIGRAM EXTRACTION

  1  It is easy for Americans to obtain a gun.
  2  It will be easy for Americans to obtain a

```
         gun.
      3  It is easy for Americans to obtain gun.
      4  It is easy for American to obtain a gun.
      5  It was easy for Americans to obtain a gun.

  TOTAL EXECUTION TIME:   23.30 CPU seconds.
```

INPUT

```
  (H /  |have the quality of being|
    :DOMAIN (H2 /  |have the quality of being|
              :DOMAIN (E /  |eat, take in|
                        :AGENT YOU
                        :PATIENT (P /  |poulet|))
              :RANGE (O /  |obligatory|))
    :RANGE (P2 /  |possible, potential|))
```

LATTICE CREATED

```
  260 nodes, 703 arcs, 10,734,304 paths;
  48 distinct unigrams, 345 distinct bigrams.
```

RANDOM EXTRACTION

```
  You may be obliged to eat that there was
  the poulet.

  An consumptions of poulet by you may be
  the requirements.

  It might be the requirement that the chicken
  are eaten by you.
```

DEFAULT EXTRACTION

```
  That the consumption of the chicken by you
  is obligatory is possible.
```

STATISTICAL BIGRAM EXTRACTION

```
  1  You may have to eat chicken.
  2  You might have to eat chicken.
  3  You may be required to eat chicken.
  4  You might be required to eat chicken.
  5  You may be obliged to eat chicken.

  TOTAL EXECUTION TIME:   58.78 CPU seconds.
```

A final (abbreviated) example comes from interlingua expressions produced by the semantic analyzer of JAPANGLOSS, involving long sentences characteristic of newspaper text. Note that although the lattice is not much larger than in the previous examples, it now encodes many more paths.

LATTICE CREATED

```
  267 nodes, 726 arcs,
  4,831,867,621,815,091,200 paths;
  67 distinct unigrams, 332 distinct bigrams.
```

RANDOM EXTRACTION

```
  Subsidiary on an Japan's of Perkin Elmer
  Co.'s hold a stocks's majority, and as for
  a beginnings, productia of an stepper and
  an dry etching devices which were applied
  for an constructia of microcircuit
  microchip was planed.
```

STATISTICAL BIGRAM EXTRACTION

```
  Perkin Elmer Co.'s Japanese subsidiary
  holds majority of stocks, and as for the
  beginning, production of steppers and dry
  etching devices that will be used to
  construct microcircuit chips are planned.
```

```
  TOTAL EXECUTION TIME:   106.28 CPU seconds.
```

## 9  Strengths and Weaknesses

Many-paths generation leads to a new style of incremental grammar building. When dealing with some new construction, we first rather mindlessly overgenerate, providing the grammar with many ways to express the same thing. Then we watch the statistical component make its selections. If the selections are correct, there is no need to refine the grammar.

For example, in our first grammar, we did not make any lexical or grammatical case distinctions. So our lattices included paths like *Him saw I* as well as *He saw me*. But the statistical model studiously avoided the bad paths, and in fact, we have yet to see an incorrect case usage from our generator. Likewise, our grammar proposes both *his box* and *the box of he/him*, but the former is statistically much more likely. Finally, we have no special rule to prohibit articles and possessives from appearing in the same noun phrase, but the bigram *the his* is so awful that the null article is always selected in the presence of a possessive pronoun. So we can get away with treating possessive pronouns like regular adjectives, greatly simplifying our grammar.

We have also been able to simplify the generation of morphological variants. While true irregular forms (e.g., *child/children*) must be kept in a small exception table, the problem of "multiple regular" patterns usually increases the size of this table dramatically. For example, there are two ways to pluralize a noun ending in *-o*, but often only one is correct for a given noun (*potatoes*, but *photos*). There are many such inflectional and derivational patterns. Our approach is to apply all patterns and insert all results into the word lattice. Fortunately, the statistical model steers clear of sentences containing nonwords like *potatos* and *photoes*. We thus get by with a very small exception table, and furthermore, our spelling habits automatically adapt to the training corpus.

Most importantly, the two-level generation model allows us to indirectly apply lexical constraints for the selection of open-class words, even though these constraints are not explicitly represented in the generator's lexicon. For example, the selection of a word from a pair of frequently co-occurring adjacent words will automatically create a strong bias for the selection of the other member of the pair, if the latter is compatible with the semantic concept being lexicalized. This type of collocational knowledge, along with additional collocational information based on long- and variable-distance dependencies, has been successfully used in the past to increase the fluency of generated text (Smadja and McKeown, 1991). But, although such collocational information can be extracted automatically, it has to be manually reformulated into the generator's representational framework before it can be used as an additional constraint during pure knowledge-based generation. In contrast, the two-level model provides for the automatic collection and implicit representation of collocational constraints between adjacent words.

In addition, in the absence of external lexical constraints the language model prefers words more typical of and common in the domain, rather than generic or overly specialized or formal alternatives. The result is text that is more fluent and closely simulates the style of the training corpus in this respect. Note for example the choice of *obtain* in the second example of the previous section in favor of the more formal *procure*.

Many times, however, the statistical model does not finish the job. A bigram model will happily select a sentence like *I only hires men who is good pilots*. If we see plenty of output like this, then grammatical work on agreement is needed. Or consider *They planned increase in production*, where the model drops an article because *planned increase* is such a frequent bigram. This is a subtle interaction—is *planned* a main verb or an adjective? Also, the model prefers short sentences to long ones with the same semantic content, which favors conciseness, but sometimes selects bad n-grams to avoid a longer (but clearer) rendition. This an interesting problem not encountered in otherwise similar speech recognition models. We are currently investigating solutions to all of these problems in a highly experimental setting.

## 10 Conclusions

Statistical methods give us a way to address a wide variety of knowledge gaps in generation. They even make it possible to load non-traditional duties onto a generator, such as word sense disambiguation for machine translation. For example, *bei* in Japanese may mean either *American* or *rice*, and *sha* may mean *shrine* or *company*. If for some reason, the analysis of *beisha* fails to resolve these ambiguities, the generator can pass them along in the lattice it builds, e.g.:

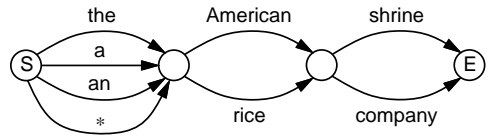

In this case, the statistical model has a strong preference for *the American company*, which is nearly always the correct translation.[7]

Furthermore, our two-level generation model can implicitly handle both paradigmatic and syntagmatic lexical constraints, leading to the simplification of the generator's grammar and lexicon, and enhancing portability. By retraining the statistical component on a different domain, we can automatically pick up the peculiarities of the sublanguage such as preferences for particular words and collocations. At the same time, we take advantage of the strength of the knowledge-based approach which guarantees grammatical inputs to the statistical component, and reduces the amount of language structure that is to be retrieved from statistics. This approach addresses the problematic aspects of both pure knowledge-based generation (where incomplete knowledge is inevitable) and pure statistical *bag generation* (Brown et al., 1993) (where the statistical system has no linguistic guidance).

Of course, the results are not perfect. We can improve on them by enhancing the statistical model, or by incorporating more knowledge and constraints in the lattices, possibly using automatic knowledge acquisition methods. One direction we intend to pursue is the rescoring of the top $N$ generated sentences by more expensive (and extensive) methods, incorporating for example stylistic features or explicit knowledge of flexible collocations.

## Acknowledgments

We would like to thank Yolanda Gil, Eduard Hovy, Kathleen McKeown, Jacques Robin, Bill Swartout, and the ACL reviewers for helpful comments on earlier versions of this paper. This work was supported in part by the Advanced Research Projects Agency (Order 8073, Contract MDA904-91-C-5224) and by the Department of Defense.

## References

John Bateman. 1990. Upper modeling: A level of semantics for natural language processing. In

---

[7]See also (Dagan and Itai, 1994) for a study of the use of lexical co-occurrences to choose among open-class word translations.

*Proc. Fifth International Workshop on Natural Language Generation*, pages 54–61, Dawson, PA.

Peter F. Brown, S. A. Della Pietra, V. J. Della Pietra, and R. L. Mercer. 1993. The mathematics of statistical machine translation: Parameter estimation. *Computational Linguistics*, 19(2), June.

Yen-Lu Chow and Richard Schwartz. 1989. The N-Best algorithm: An efficient search procedure for finding top N sentence hypotheses. In *Proc. DARPA Speech and Natural Language Workshop*, pages 199–202.

Kenneth W. Church and William A. Gale. 1991. A comparison of the enhanced Good-Turing and deleted estimation methods for estimating probabilities of English bigrams. *Computer Speech and Language*, 5:19–54.

Ido Dagan and Alon Itai. 1994. Word sense disambiguation using a second language monolingual corpus. *Computational Linguistics*, 20(4):563–596.

Laurence Danlos. 1986. *The Linguistic Basis of Text Generation*. Studies in Natural Language Processing. Cambridge University Press.

Michael Elhadad and Jacques Robin. 1992. Controlling content realization with functional unification grammars. In Robert Dale, Eduard Hovy, Dietmar Rösner, and Oliviero Stock, editors, *Aspects of Automated Natural Language Generation*, pages 89–104. Springler Verlag.

Michael Elhadad. 1993. *Using Argumentation to Control Lexical Choice: A Unification-Based Implementation*. Ph.D. thesis, Computer Science Department, Columbia University, New York.

Karin Harbusch, Gen-ichiro Kikui, and Anne Kilger. 1994. Default handling in incremental generation. In *Proc. COLING-94*, pages 356–362, Kyoto, Japan.

Vasileios Hatzivassiloglou and Kevin Knight. 1995. Unification-based glossing. In *Proc. IJCAI*.

Philip J. Hayes. 1981. A construction-specific approach to focused interaction in flexible parsing. In *Proc. ACL*, pages 149–152.

Annette Herskovits. 1986. *Language and spatial cognition: an interdisciplinary study of the prepositions of English*. Studies in Natural Language Processing. Cambridge University Press.

Kevin Knight and Ishwar Chander. 1994. Automated postediting of documents. In *Proc. AAAI*.

Kevin Knight and Steve K. Luk. 1994. Building a large-scale knowledge base for machine translation. In *Proc. AAAI*.

Kevin Knight, Ishwar Chander, Matthew Haines, Vasileios Hatzivassiloglou, Eduard Hovy, Masayo Iida, Steve K. Luk, Akitoshi Okumura, Richard Whitney, and Kenji Yamada. 1994. Integrating knowledge bases and statistics in MT. In *Proc. Conference of the Association for Machine Translation in the Americas (AMTA)*.

Kevin Knight, Ishwar Chander, Matthew Haines, Vasileios Hatzivassiloglou, Eduard Hovy, Masayo Iida, Steve K. Luk, Richard Whitney, and Kenji Yamada. 1995. Filling knowledge gaps in a broad-coverage MT system. In *Proc. IJCAI*.

Karen Kukich, K. McKeown, J. Shaw, J. Robin, N. Morgan, and J. Phillips. 1994. User-needs analysis and design methodology for an automated document generator. In A. Zampolli, N. Calzolari, and M. Palmer, editors, *Current Issues in Computational Linguistics: In Honour of Don Walker*. Kluwer Academic Press, Boston.

Karen Kukich. 1988. Fluency in natural language reports. In David D. McDonald and Leonard Bolc, editors, *Natural Language Generation Systems*. Springer-Verlag, Berlin.

Alon Lavie. 1994. An integrated heuristic scheme for partial parse evaluation. In *Proc. ACL (student session)*.

Marie W. Meteer, D. D. McDonald, S. D. Anderson, D. Forster, L. S. Gay, A. K. Huettner, and P. Sibun. 1987. Mumble-86: Design and implementation. Technical Report COINS 87-87, University of Massachussets at Amherst, Ahmerst, MA.

George A. Miller. 1990. Wordnet: An on-line lexical database. *International Journal of Lexicography*, 3(4). (Special Issue).

Penman. 1989. The Penman documentation. Technical report, USC/Information Sciences Institute.

Paul Procter, editor. 1978. *Longman Dictionary of Contemporary English*. Longman, Essex, UK.

Jacques Robin. 1995. *Revision-Based Generation of Natural Language Summaries Providing Historical Background: Corpus-based Analysis, Design, Implementation, and Evaluation*. Ph.D. thesis, Computer Science Department, Columbia University, New York, NY. Also, Technical Report CU-CS-034-94.

Frank Smadja and Kathleen R. McKeown. 1991. Using collocations for language generation. *Computational Intelligence*, 7(4):229–239, December.

M. Tomita and E. Nyberg. 1988. The GenKit and Transformation Kit User's Guide. Technical Report CMU-CMT-88-MEMO, Center for Machine Translation, Carnegie Mellon University.

A. Waibel and K. F. Lee, editors. 1990. *Readings in Speech Recognition*. Morgan Kaufmann, San Mateo, CA.

R. Weischedel and J. Black. 1980. Responding to potentially unparseable sentences. *Am. J. Computational Linguistics*, 6.